\documentclass[runningheads]{llncs}
\usepackage[T1]{fontenc}
\usepackage{booktabs}
\usepackage{bm}
\usepackage{amsmath}
\usepackage{amssymb}
\usepackage{multirow}
\usepackage[table]{xcolor} 
\usepackage{subcaption}
\usepackage{hyperref}
\usepackage{bbding}

\usepackage{graphicx,verbatim}
\begin{document}
\title{AEM: Attention Entropy Maximization	for Multiple Instance
	Learning based Whole Slide Image Classification}
%
\begin{comment}  %% Removed for anonymized MICCAI 2025 submission
\author{First Author\inst{1}\orcidID{0000-1111-2222-3333} \and
Second Author\inst{2,3}\orcidID{1111-2222-3333-4444} \and
Third Author\inst{3}\orcidID{2222--3333-4444-5555}}
%
\authorrunning{F. Author et al.}
\institute{Princeton University, Princeton NJ 08544, USA \and
Springer Heidelberg, Tiergartenstr. 17, 69121 Heidelberg, Germany
\email{lncs@springer.com}\\
\url{http://www.springer.com/gp/computer-science/lncs} \and
ABC Institute, Rupert-Karls-University Heidelberg, Heidelberg, Germany\\
\email{\{abc,lncs\}@uni-heidelberg.de}}

\end{comment}

	\author{
		Yunlong Zhang\inst{1,2} \and
		Honglin Li\inst{1,2} \and
		Yuxuan Sun\inst{1,2} \and
		Zhongyi Shui\inst{1,2} \and
		Jingxiong Li\inst{1,2} \and
		Chenglu Zhu\inst{2} \and
		Lin Yang\inst{2(}\Envelope\inst{)}
	}
	
	\institute{College of Computer Science and Technology, Zhejiang University \and
		School of Engineering, Westlake University \\
		yanglin@westlake.edu.cn
	}

\maketitle              % typeset the header of the contribution

\begin{abstract}
Multiple Instance Learning (MIL) effectively analyzes whole slide images but faces overfitting due to attention over-concentration. While existing solutions rely on complex architectural modifications or additional processing steps, we introduce Attention Entropy Maximization (AEM), a simple yet effective regularization technique. Our investigation reveals the positive correlation between attention entropy and model performance. Building on this insight, we integrate AEM regularization into the MIL framework to penalize excessive attention concentration.  To address sensitivity to the AEM weight parameter, we implement Cosine Weight Annealing, reducing parameter dependency. Extensive evaluations demonstrate AEM's superior performance across diverse feature extractors, MIL frameworks, attention mechanisms, and augmentation techniques. Here is our anonymous code: \url{https://github.com/dazhangyu123/AEM}.

% Authors must provide keywords and are not allowed to remove this Keyword section.

\end{abstract}
\section{Introduction}

Whole slide images (WSIs) are widely recognized as the gold standard for numerous cancer diagnoses, playing a crucial role in ensuring precise diagnosis \cite{bejnordi2017diagnostic}, prognosis \cite{yao2020whole}, and the development of treatment plans \cite{song2023artificial}. In recent years, attention-based multiple instance learning (ABMIL) \cite{ilse2018attention} has emerged as a promising approach for WSI analysis. However, recent studies have uncovered overfitting issues in MIL due to factors like limited available data \cite{tang2023multiple,zhang2022dtfd,zhang2023attention,li2021dual}, class imbalance \cite{zhang2023attention}, and staining bias \cite{lin2023interventional,zhang2022benchmarking}.

In the attention mechanism, attention values represent the importance or relevance of instances to the bag prediction, influencing both prediction accuracy and result interpretability. Relevant studies \cite{yufei2022bayes,zhang2023attention} have revealed that excessive concentration of attention values in ABMIL hinders model interpretability and results in overfitting \cite{zhang2023attention}. There have been several solutions for alleviating attention concentration. Masking-based methods \cite{qu2022bi,tang2023multiple,zhang2023attention} mask out the instances with the highest attention values, allocating their attention values to remaining instances. Clustering-based methods \cite{sharma2021cluster,guan2022node} group instances into clusters and randomly sample instances from these clusters, ensuring attention values are not overly focused on minority instances. ACMIL \cite{zhang2023attention} generates the heatmap by averaging the attention values generated by multiple attention heads, thereby avoiding the over-concentration of attention values. DGR-MIL \cite{bai2025norma} addresses this through learnable global vectors capturing diverse patterns via cross-attention, with strategies to push vectors toward positive instance centers and enforce orthogonality using DPP-based diversity loss. However, most solutions add complexity and computational overhead, limiting flexibility (see Table \ref{tab:compare}).

\begin{table}[tbp]
	\caption{Comparison of WSI classification methods addressing overfitting.}\label{tab:compare}
	\centering
	\scalebox{1.0}{
		\resizebox{0.8\columnwidth}{!}{%
			\begin{tabular}{l|c}
				\toprule
				\textbf{Method}  & \textbf{Extra Modules/Processing} \\
				\midrule
				DTFD-MIL~\cite{zhang2022dtfd} & Double-tier attention mechanisms \\
				IBMIL~\cite{lin2023interventional} & New training stage of interventional training from scratch\\
				C2C~\cite{sharma2021cluster} & Clustering and sampling process \\
				MHIM-MIL~\cite{tang2023multiple} & Teacher model for masking easy instances \\
				ACMIL~\cite{zhang2023attention}& Multiple branch attention for extracting pattern embeddings  \\
				DGR-MIL \cite{bai2025norma} & Instance center pushing and DPP-based vector orthogonality \\
				\midrule
				\textbf{AEM(ours)}& \textbf{None} \\
				\bottomrule
	\end{tabular}}}
\end{table}

\begin{figure*}[!t]
	\centering
	\includegraphics[width=\textwidth]{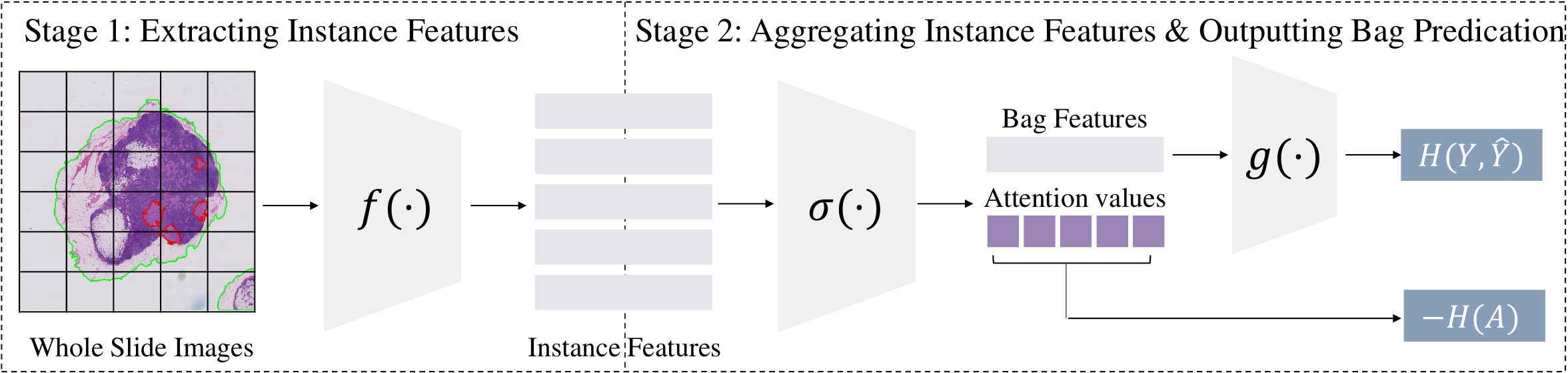}
	%\includesvg[width=0.92\textwidth]{image/overview3.svg}
	\caption{Overview of plugging AEM into MIL framework. AEM adds only a negative entropy regularization for attention values to the regular MIL framework.}
	\label{fig_overview}
\end{figure*}

To address the limitations of existing complex solutions, we propose Attention Entropy Maximization (AEM), a lightweight yet powerful approach for mitigating attention concentration and MIL method overfitting. Our empirical analysis establishes a positive correlation between attention entropy and model performance, which forms the foundation for developing AEM. The approach integrates a negative entropy loss term for attention values into the standard MIL framework (Figure \ref{fig_overview}), promoting a more uniform distribution of attention across instances.
To address sensitivity to the AEM weight parameter, we introduce Cosine Weight Annealing, reducing parameter dependency. Unlike existing overfitting mitigation techniques, AEM requires no additional modules or processing steps, enabling seamless integration with current MIL frameworks while maintaining computational efficiency.

Our experimental evaluations on three datasets (CAMELYON16, CAMELYON17, and our in-house LBC dataset) demonstrate AEM's superior performance over existing methods. Furthermore, extensive experiments showcase AEM's versatility, effectively combining with five feature extractors (Lunit pretrained ViT-S \cite{kang2023benchmarking},  PathGen-CLIP pretrained ViT-L \cite{sun2024pathgen16m16millionpathology}, UNI pretrained ViT-L \cite{chen2024towards}, CONCH pretrained ViT-B \cite{lu2024visual}, and GigaPath pretrained ViT-G \cite{xu2024whole}), Subsampling augmentation technique, two advanced MIL frameworks (DTFD-MIL \cite{zhang2022dtfd} and ACMIL \cite{zhang2023attention}), and three attention mechanisms (DSMIL \cite{li2021dual}, LongNet \cite{ding2023longnet} and MHA \cite{vaswani2017attention}). These results underscore AEM's potential as a widely applicable enhancement to existing MIL methodologies in medical image analysis.

\section{Method}

% We present AEM, a versatile regularization technique that enhances MIL frameworks for WSI classification. To establish AEM's core principles, we introduce it in conjunction with ABMIL, a fundamental and widely adopted framework for processing WSI data. This methodological foundation not only simplifies the exposition but also demonstrates how AEM's mechanisms can be readily extended to more sophisticated MIL variants. This section first provides a comprehensive overview of ABMIL for WSI classification (Section \ref{sec:abmil}), followed by a detailed exposition of AEM's motivation and implementation (Section \ref{sec:AEM}), and concludes with an exploration of adaptive weight scheduling strategies to optimize AEM's performance.

\subsection{ABMIL for WSI Analysis} \label{sec:abmil}
\noindent\textbf{MIL formulation.} For WSI classification, we have the WSI $\bm X$ with slide-level label $\bm Y$. Due to the extreme resolution of WSIs ($50,000\times50,000$ to $100,000\times100,000$), direct training is computationally infeasible. ABMIL \cite{ilse2018attention} addresses this by segmenting WSIs into non-overlapping patches $\{\bm x_n\}_{n=1}^N$ and employing a two-step process to predict the slide label $\hat{\bm Y}$.

\noindent\textbf{Extracting instance features.} Current MIL methods typically use features from a frozen backbone like ImageNet-pretrained ResNet. Recent studies \cite{lu2021data,dehaene2020self} show that using encoders pre-trained with self-supervised learning and vision-language pretraining improves performance. To comprehensively verify AEM's effectiveness, we use five feature extractors: DINO pretrained ViT-S/16 \cite{kang2023benchmarking}, PathGen-CLIP pretrained ViT-L/14 \cite{sun2024pathgen16m16millionpathology}, UNI pretrained ViT-L \cite{chen2024towards}, CONCH pretrained ViT-B \cite{lu2024visual}, and GigaPath pretrained ViT-G \cite{xu2024whole}).

\noindent\textbf{Aggregating instance features and outputting bag predication.} ABMIL aggregates instance embeddings into the bag embedding using a gated attention operator:
\begin{equation} \label{eq2}
\bm{z} = \sum_{n=1}^{N} a_n \bm{h}_{n},
\end{equation}
where $a_n = \sigma (\bm{h}_{n})$ represents the attention values for the $n$-th instance, $\bm{h}_{n}$. The bag prediction is then obtained through an MLP layer: $\hat{\bm Y} = g(\bm z)$. 
% The training loss is defined as:
% \begin{equation} \label{ce_loss}
% L_{ce} = H(\bm Y, \hat{\bm Y}) = -\sum_{i} \bm Y_i \log \hat{\bm Y}_i.
% \end{equation}

While we initially demonstrate our approach with ABMIL, our proposed AEM method (detailed in Section \ref{sec:AEM}) can be effectively applied to various attention mechanisms, including DSMIL \cite{li2021dual}, MHA \cite{zhang2023attention}, and LongNet \cite{xu2024whole}.

\subsection{Attention Entropy Maximization} \label{sec:AEM}
\begin{figure}[t]
	\centering
	\includegraphics[width=0.5\linewidth]{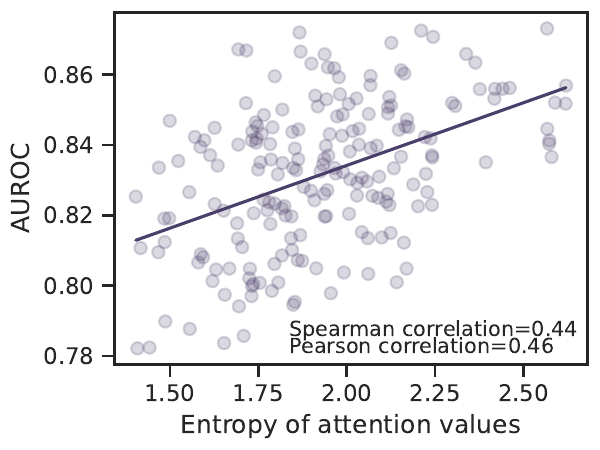}
	\caption{%
		There exists a positive correlation between  AUROC values and entropy of attention values across experimental seeds.  One point denotes the outcome of a single seed on the LBC dataset.
	}
	\label{fig:auc-entropy}
    \vspace{-3mm}
\end{figure}

\noindent\textbf{Motivation.}
Studies show that low attention entropy can cause training instability and poor generalization in attention-based models \cite{zhai2023stabilizing,xiong2020layer,dehghani2023scaling}. To investigate this in WSI classification, we trained ABMIL with 200 different random initializations while keeping training, validation, and test sets fixed. Figure \ref{fig:auc-entropy} reveals a positive correlation between AUROC performance and attention entropy values on the test set, with higher entropy consistently associated with better classification results. These findings highlight the importance of attention diversity for effective WSI analysis, demonstrating that maintaining high attention entropy improves model performance and generalization.

\noindent\textbf{Implementation.} AEM maximizes the entropy $H(A)$ of attention values, $A = \{a_n\}_{n=1}^{N}$, by formulating it as negative entropy \cite{shannon1948mathematical}:

\begin{equation} \label{AEM}
L_{aem} = -H(A) = \sum_{n} a_n \log a_n.
\end{equation}

This encourages consideration of more informative regions in WSIs, potentially improving generalization. The final objective is formulated as:

\begin{equation} \label{all_loss}
L_{total} = L_{ce} + \lambda L_{aem},
\end{equation}
where $\lambda$ balances the trade-off between losses. The weight parameter $\lambda$ in AEM balances attention focusing and entropy regularization. Too small values lead to insufficient instance diversity, while too large values force uniform attention distribution, reducing to mean-pooling behavior—consistently shown inferior to attention-based MIL approaches \cite{ilse2018attention}.

We adopt Cosine Weight Annealing \cite{loshchilov2017decoupled} as our scheduling strategy, which gradually reduces $\lambda$ following a cosine curve. This approach naturally supports AEM's progression: initially maintaining high entropy for broad instance exploration when features are less reliable, then transitioning to focused attention as discriminative capabilities improve. 

\noindent\textbf{Discussion.} AEM serves a similar role to the KL-divergence loss in C2C \cite{sharma2021cluster} by promoting attention distribution, but with key differences. AEM operates globally across all instances, while C2C's KL-divergence works only within individual clusters. Unlike C2C's strict uniform enforcement, AEM's negative entropy approach provides flexibility by penalizing extreme concentration while allowing meaningful non-uniform distributions when appropriate \cite{han2021diversity}. Our experiments confirm that replacing AEM with KL-divergence decreases performance.

%\subsection{Further exploration} \label{sec:exploration}
%\noindent\textbf{Normalized Attention Entropy.} In WSI analysis tasks, where instance numbers typically range from hundreds to tens of thousands, the conventional attention entropy $H(A)$ becomes scale-dependent and fails to provide a consistent measure of attention concentration across varying instance counts. To address this limitation, we introduce the Normalized Attention Entropy measure $H_N(A)$, defined as:
%\begin{equation} \label{normalized_AEM}
%	H_N(A) = -\frac{1}{\log N} \sum_{n} a_n \log a_n,
%\end{equation}
%where $N$ denotes the total number of instances and $a_n$ represents the attention weight for the $n$-th instance. This normalization by $\log N$ bounds the entropy measure to $[0,1]$, enabling fair comparisons across datasets with varying instance counts while preserving the relative attention distribution characteristics.

% \subsection{Cosine Weight Annealing}

\section{Experiments} \label{sec:expriment}

\begin{table*}[!t]
	\centering
    \tiny  % Reduce font size to better fit content
    \setlength{\tabcolsep}{4pt}  % Adjust column spacing
	\caption{The performance of different MIL approaches across three datasets and two evaluation metrics.  The most superior performance is highlighted in \textbf{bold}.  }\label{tab:main}
	
	\begin{tabular}{@{}l|*{6}{c}@{}} 
		\toprule
		\multirow{2}{*}{Method} & \multicolumn{2}{c}{CAMELYON-16} & \multicolumn{2}{c}{CAMELYON-17} &  \multicolumn{2}{c}{LBC} \\ 
		\cmidrule(lr){2-3} \cmidrule(lr){4-5} \cmidrule(lr){6-7}
		& F1-score       & AUC           & F1-score       & AUC       & F1-score            & AUC   \\\midrule
				\multicolumn{7}{c}{SSL pretrained ViT-S (Lunit \cite{kang2023benchmarking})} \\
                \midrule

		Clam-SB \cite{lu2021data} & 0.925\tiny{$\pm$0.035} & 0.969\tiny{$\pm$0.024} & 0.523\tiny{$\pm$0.020} & 0.846\tiny{$\pm$0.020} & 0.617\tiny{$\pm$0.022} & 0.865\tiny{$\pm$0.018}  \\ 
		LossAttn \cite{shi2020loss}                   
		& {0.908\tiny{$\pm$0.031}}             & {0.928\tiny{$\pm$0.014}}             & 0.575\tiny{$\pm$0.051} &	0.865\tiny{$\pm$0.016}
		& {0.621\tiny{$\pm$0.012}} &	{0.843\tiny{$\pm$0.006}}   \\
		TransMIL \cite{shao2021transmil} & 0.922\tiny{$\pm$0.019} & 0.943\tiny{$\pm$0.009} & 0.554\tiny{$\pm$0.048} & 0.792\tiny{$\pm$0.029} & 0.539\tiny{$\pm$0.028} & 0.805\tiny{$\pm$0.010}  \\ 
		DSMIL \cite{li2021dual} & 0.943\tiny{$\pm$0.007} & 0.966\tiny{$\pm$0.009} & 0.532\tiny{$\pm$0.064} & 0.804\tiny{$\pm$0.032} & 0.562\tiny{$\pm$0.028} & 0.820\tiny{$\pm$0.033} \\

		IBMIL \cite{lin2023interventional} & 0.912\tiny{$\pm$0.034} &	0.954\tiny{$\pm$0.022} &	{0.557\tiny{$\pm$0.034}} &	{0.850\tiny{$\pm$0.024}} &	0.604\tiny{$\pm$0.032} &	0.834\tiny{$\pm$0.014} \\ 
		MHIM-MIL  \cite{tang2023multiple}                     
		&0.932\tiny{$\pm$0.024}&	{0.970\tiny{$\pm$0.037}}&	0.541\tiny{$\pm$0.022}&	0.845\tiny{$\pm$0.026}&	{0.658\tiny{$\pm$0.041}}&	{0.872\tiny{$\pm$0.022}}

		\\
		ILRA  \cite{xiang2022exploring}                  
		&0.904\tiny{$\pm$0.071} &	0.940\tiny{$\pm$0.060} &	{0.631\tiny{$\pm$0.051}} &	0.860\tiny{$\pm$0.020} &	0.618\tiny{$\pm$0.051} &	0.859\tiny{$\pm$0.017} 
		\\
		ABMIL \cite{ilse2018attention}                
		&0.914\tiny{$\pm$0.031} &	0.945\tiny{$\pm$0.027} &	{0.522\tiny{$\pm$0.050}} &	0.853\tiny{$\pm$0.016} &	0.595\tiny{$\pm$0.036} &	0.831\tiny{$\pm$0.022}  						\\ 
		\rowcolor{gray!20} AEM(ours)  &\textbf{0.947\tiny{$\pm$0.003}}&\textbf{0.974\tiny{$\pm$0.007}}&\textbf{0.647\tiny{$\pm$0.007}}&\textbf{0.887\tiny{$\pm$0.013}}&\textbf{0.664\tiny{$\pm$0.021}}&\textbf{0.879\tiny{$\pm$0.013}} \\
		\midrule
		\multicolumn{7}{c}{VLM pretrained ViT-L (PathGen-CLIP \cite{sun2024pathgen16m16millionpathology})} \\
		\midrule

		Clam-SB \cite{lu2021data} & 0.941\tiny{$\pm$0.014} & 0.960\tiny{$\pm$0.015} & 0.622\tiny{$\pm$0.031} & 0.899\tiny{$\pm$0.012} & 0.641\tiny{$\pm$0.025} & 0.870\tiny{$\pm$0.013} \\
		LossAttn \cite{shi2020loss}  & 0.948\tiny{$\pm$0.004} & 0.981\tiny{$\pm$0.017} & 0.667\tiny{$\pm$0.023} & 0.891\tiny{$\pm$0.009} & 0.657\tiny{$\pm$0.035} & 0.874\tiny{$\pm$0.006} \\
		TransMIL \cite{shao2021transmil} & 0.951\tiny{$\pm$0.024} & 0.968\tiny{$\pm$0.028} & 0.656\tiny{$\pm$0.021} & 0.892\tiny{$\pm$0.014} & 0.573\tiny{$\pm$0.019} & 0.849\tiny{$\pm$0.010} \\
		DSMIL \cite{li2021dual} & 0.895\tiny{$\pm$0.038} & 0.949\tiny{$\pm$0.017} & 0.582\tiny{$\pm$0.062} & 0.887\tiny{$\pm$0.013} & 0.586\tiny{$\pm$0.024} & 0.848\tiny{$\pm$0.010} \\
		IBMIL \cite{lin2023interventional} & 0.935\tiny{$\pm$0.014} & 0.953\tiny{$\pm$0.009} & 0.629\tiny{$\pm$0.027} & 0.884\tiny{$\pm$0.016} & 0.640\tiny{$\pm$0.010} & 0.867\tiny{$\pm$0.007} \\
		MHIM-MIL \cite{tang2023multiple}   & 0.946\tiny{$\pm$0.033} & 0.984\tiny{$\pm$0.016} & 0.594\tiny{$\pm$0.090} & 0.912\tiny{$\pm$0.009} & 0.660\tiny{$\pm$0.030} & \textbf{0.890\tiny{$\pm$0.007}} \\
		ILRA \cite{xiang2022exploring}  & 0.929\tiny{$\pm$0.018} & 0.963\tiny{$\pm$0.019} & 0.662\tiny{$\pm$0.048} & \textbf{0.914\tiny{$\pm$0.017}} & 0.626\tiny{$\pm$0.028} & 0.864\tiny{$\pm$0.014} \\
		ABMIL \cite{ilse2018attention}  & 0.953\tiny{$\pm$0.018} & 0.972\tiny{$\pm$0.010} & 0.610\tiny{$\pm$0.025} & 0.864\tiny{$\pm$0.017} & 0.621\tiny{$\pm$0.023} & 0.853\tiny{$\pm$0.013} \\
		\rowcolor{gray!20} AEM(ours)  &\textbf{0.967\tiny{$\pm$0.025}}&\textbf{0.988\tiny{$\pm$0.013}}&\textbf{0.688\tiny{$\pm$0.016}}&{0.905\tiny{$\pm$0.005}}&\textbf{0.691\tiny{$\pm$0.032}}&{0.884\tiny{$\pm$0.010}} \\
        \midrule

			\multicolumn{7}{c}{SSL pretrained ViT-L	 (UNI \cite{chen2024towards})} \\
			\midrule
			
			ABMIL \cite{ilse2018attention}  & 0.968\tiny{$\pm$0.011} & {0.996\tiny{$\pm$0.003}} & 0.605\tiny{$\pm$0.047} & \textbf{0.885\tiny{$\pm$0.015}} & 0.580\tiny{$\pm$0.023} & 0.844\tiny{$\pm$0.024} \\
			\rowcolor{gray!20} AEM(ours)  &\textbf{0.975\tiny{$\pm$0.003}}&\textbf{0.998\tiny{$\pm$0.003}}&\textbf{0.633\tiny{$\pm$0.024}}&{0.863\tiny{$\pm$0.017}}&\textbf{0.645\tiny{$\pm$0.021}}&\textbf{0.870\tiny{$\pm$0.015}} \\
			\midrule
			\multicolumn{7}{c}{SSL pretrained ViT-G	 (GigaPath \cite{xu2024whole})} \\
			\midrule

			ABMIL \cite{ilse2018attention}  & {0.978\tiny{$\pm$0.007}} & \textbf{0.984\tiny{$\pm$0.009}} & 0.555\tiny{$\pm$0.040} & 0.880\tiny{$\pm$0.023} & 0.623\tiny{$\pm$0.023} & 0.866\tiny{$\pm$0.014} \\
			\rowcolor{gray!20} AEM(ours)  &\textbf{0.981\tiny{$\pm$0.009}}&0.982\tiny{$\pm$0.011}&\textbf{0.571\tiny{$\pm$0.029}}&\textbf{0.886\tiny{$\pm$0.014}}&\textbf{0.663\tiny{$\pm$0.017}}&\textbf{0.903\tiny{$\pm$0.014}} \\
			\midrule
			\multicolumn{7}{c}{VLM pretrained ViT-B	 (CONCH \cite{lu2024visual})} \\
			\midrule

			ABMIL \cite{ilse2018attention}  & 0.932\tiny{$\pm$0.015} & {0.952\tiny{$\pm$0.017}} & 0.529\tiny{$\pm$0.022} & 0.862\tiny{$\pm$0.014} & 0.589\tiny{$\pm$0.036} & 0.849\tiny{$\pm$0.023} \\
			\rowcolor{gray!20} AEM(ours)  &\textbf{0.942\tiny{$\pm$0.011}}&\textbf{0.961\tiny{$\pm$0.016}}&\textbf{0.581\tiny{$\pm$0.013}}&\textbf{0.893\tiny{$\pm$0.010}}&\textbf{0.656\tiny{$\pm$0.022}}&\textbf{0.889\tiny{$\pm$0.011}} \\
		\bottomrule
	\end{tabular}
\end{table*}

\subsection{Experimental setup}
\noindent\textbf{Datasets.} We evaluate AEM on three WSI datasets: CAMELYON16 (C16) \cite{bejnordi2017diagnostic}, CAMELYON17 (C17) \cite{bejnordi2017diagnostic}, and LBC. C16 contains 270 training WSIs from hospital 1 (split 9:1 for training/validation) and 130 testing WSIs from hospital 2. For C17, we use 500 WSIs in total, with 300 WSIs from three hospitals for training/validation (split 9:1) and 200 WSIs from two other hospitals for testing to evaluate OOD performance. The LBC dataset includes 1,989 WSIs of cervical cancer across four cytological categories: Negative, ASC-US, LSIL, and ASC-H/HSIL, split into 6:2:2 ratios for training, validation, and testing respectively.

\noindent\textbf{Implementation Details.} Following \cite{lu2021data}, we process WSIs by extracting $256 \times 256$ patches at $\times20$ magnification. The model architecture consists of a feature dimension reduction layer, gated attention network, and prediction layer, optimized using Adam with cosine learning rate decay. Hyperparameter selection was based on validation performance optimization, with default $\lambda$ values of 0.001, 0.1, and 0.2 for C16, C17, and LBC respectively. We report macro-AUC and macro-F1 scores averaged over five runs with different random initializations.

\subsection{Main results}
\noindent\textbf{AEM's effectiveness across different feature extractors.} Table \ref{tab:main} evaluates MIL approaches across three datasets using five backbones. For Lunit-pretrained ViT-S and PathGen-CLIP-pretrained ViT-L, we compare AEM against several advanced MIL methods, with our approach achieving superior performance in 10 out of 12 metrics. With ViT-S, AEM leads across all metrics, while with ViT-L, it dominates in all F1-scores and C16 AUC, with only slight trails in C17 and LBC AUC. For the remaining three backbones (UNI, GigaPath, CONCH), AEM outperforms ABMIL in 16 out of 18 metrics, demonstrating significant improvements across diverse architectures and pretraining strategies. These consistent results confirm AEM's effectiveness as a versatile enhancement applicable to various feature extractors.

\begin{figure*}[t]
	\centering
	\begin{subfigure}[b]{0.32\textwidth}
		\includegraphics[width=\textwidth]{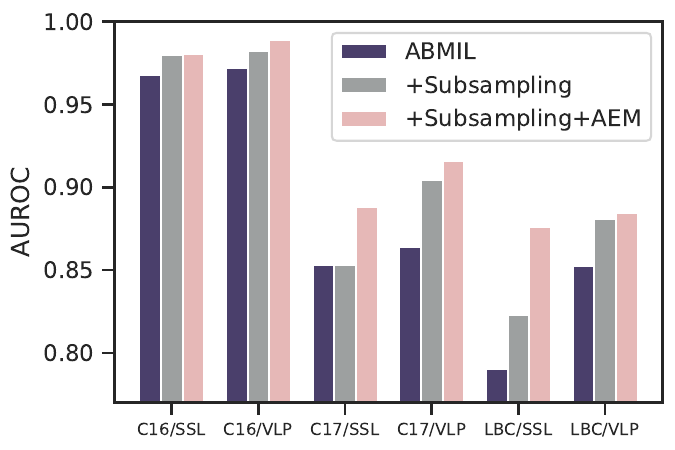}
		\caption{Subsampling}
		\label{fig:subsampling}
	\end{subfigure}
	\begin{subfigure}[b]{0.32\textwidth}
		\includegraphics[width=\textwidth]{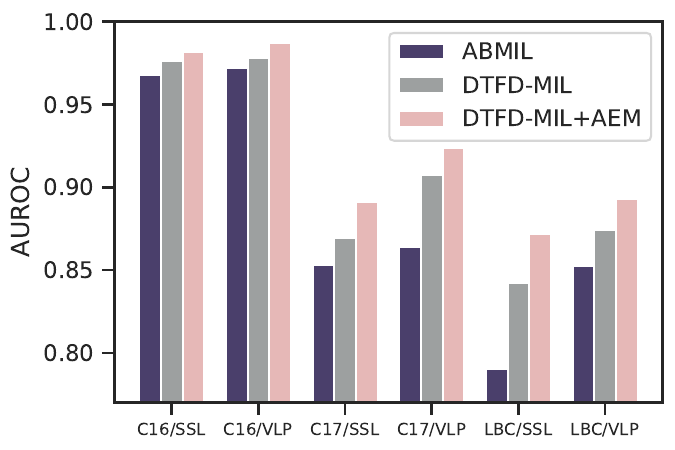}
		\caption{DTFD-MIL}
		\label{fig:dtfd}
	\end{subfigure}
	\begin{subfigure}[b]{0.32\textwidth}
		\includegraphics[width=\textwidth]{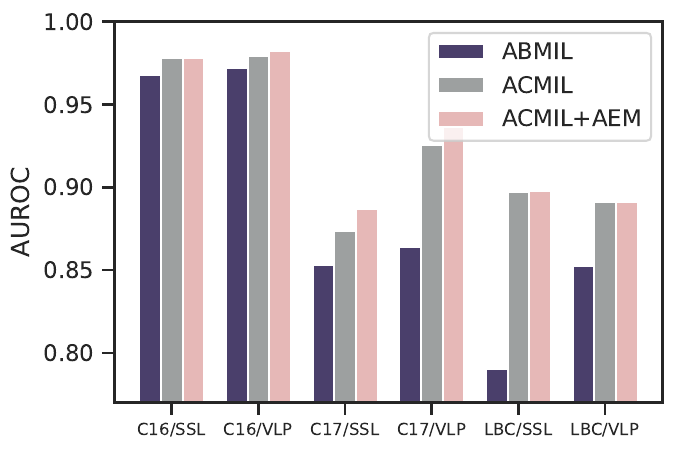}
		\caption{ACMIL}
		\label{fig:acmil}
	\end{subfigure}
	\caption{Performance comparison before and after plugging AEM into the Subsampling augmentation (a) and two advanced MIL frameworks, DTFD-MIL (b) and ACMIL (c).  C17/SSL indicates results on C17 using an SSL-pretrained backbone. \textit{AEM improves their performance on 17 out of 18 terms.}}\label{fig:adv}
\end{figure*}

\begin{figure*}[t]
	\centering
	\begin{subfigure}[b]{0.32\textwidth}
		\includegraphics[width=\textwidth]{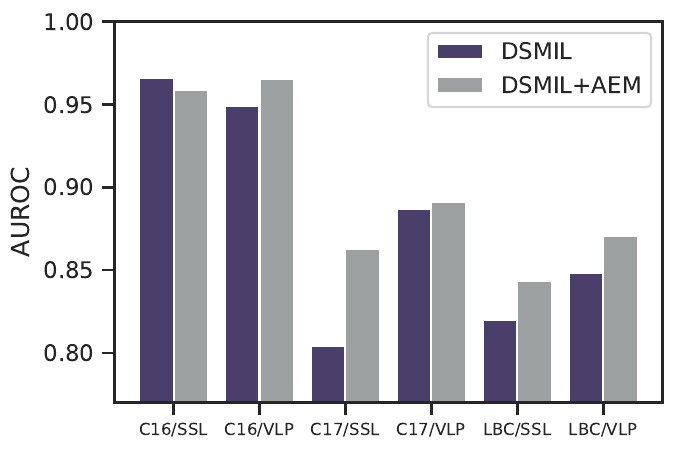}
		\caption{DSMIL}
		\label{fig:dsmil}
	\end{subfigure}
	\begin{subfigure}[b]{0.32\textwidth}
		\includegraphics[width=\textwidth]{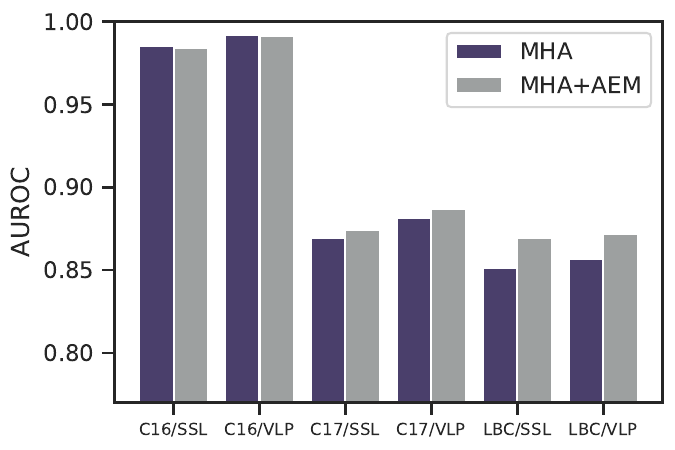}
		\caption{MHA}
		\label{fig:mha}
	\end{subfigure}
	\begin{subfigure}[b]{0.32\textwidth}
		\includegraphics[width=\textwidth]{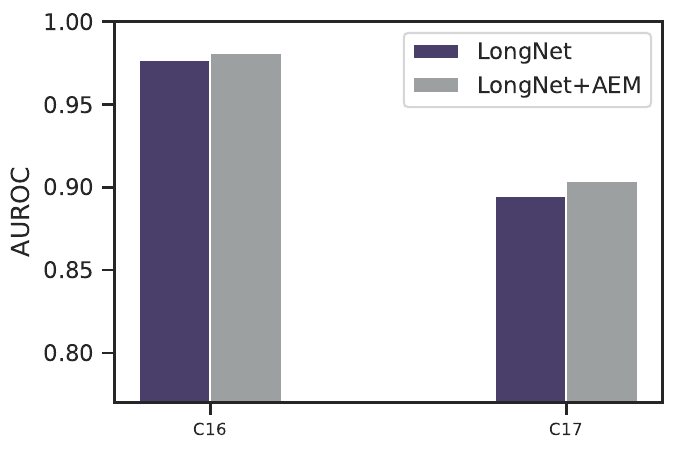}
		\caption{LongNet}
		\label{fig:longnet}
	\end{subfigure}
	\caption{Performance comparison before and after plugging AEM into DSMIL, MHA, and LongNet attention mechanisms. For LongNet, we used the pretrained Gigapath checkpoint. \textit{AEM improves performance on 11 out of 14 terms.}}\label{fig:com}
\end{figure*}

\begin{figure*}[ht]
	\centering
	\begin{subfigure}[b]{0.32\textwidth}
		\includegraphics[width=\textwidth]{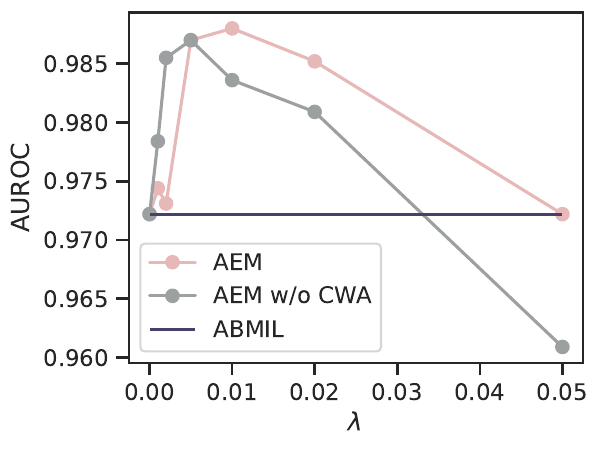}
		\caption{CAMELYON16}
		\label{fig11}
	\end{subfigure}
	\begin{subfigure}[b]{0.32\textwidth}
		\includegraphics[width=\textwidth]{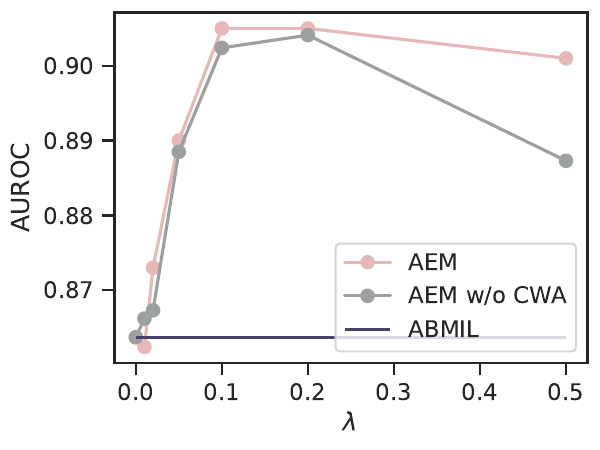}
		\caption{CAMELYON17}
		\label{fig12}
	\end{subfigure}
	\begin{subfigure}[b]{0.32\textwidth}
		\includegraphics[width=\textwidth]{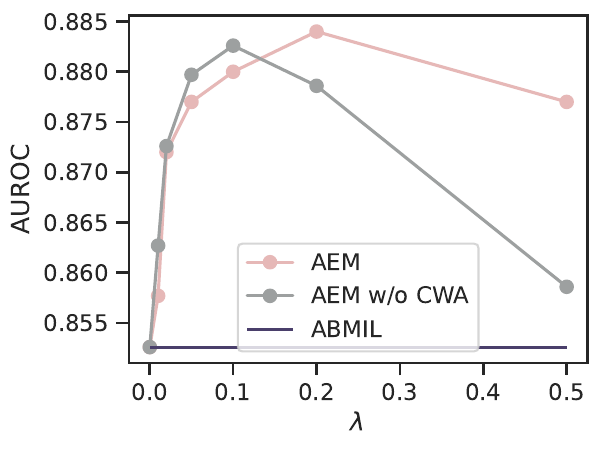}
		\caption{LBC}
		\label{fig13}
	\end{subfigure}
	\caption{Sensitivity analysis for hyperparameter $\lambda$. \textit{Choosing an appropriate $\lambda$ is critical for AEM. Moreover, including CWA can improve the stability of AEM.}}\label{fig:sensitivity}
\end{figure*}

%\begin{figure*}[ht]
%	\centering
%	\includegraphics[width=\textwidth]{figs/subsampling_comparison.pdf}
%	%\includesvg[width=0.92\textwidth]{image/overview3.svg}
%	\caption{Performance comparison of combining AEM and Subsampling augmentation. AEM can effectively improve performance upon Subsampling.}
%	\label{fig:subsampling}
%\end{figure*}

% \noindent\textbf{AEM's effectiveness across different feature extractors.} Table \ref{tab:main} demonstrates that our AEM consistently enhances the performance of ABMIL using both SSL pretrained ViT-S \cite{kang2023benchmarking} and VLM pretrained ViT-L \cite{sun2024pathgen16m16millionpathology}. To validate the method's broader applicability, we conducted additional experiments using diverse architectures and pretraining strategies. As detailed in the Supplementary Materials, AEM achieves comparable performance gains when applied to ImageNet pretrained ResNet18 \cite{he2016deep}, PathGen-CLIP pretrained ViT-B \cite{sun2024pathgen16m16millionpathology}, UNI pretrained ViT-L \cite{chen2024towards}, CONCH pretrained ViT-B \cite{lu2024visual}, and GigaPath pretrained ViT-G \cite{xu2024whole}, demonstrating its versatility across backbone networks.

\noindent\textbf{AEM enhances Subsampling, DTFD-MIL, and ACMIL.}
Figure \ref{fig:subsampling} demonstrates AEM's ability to consistently boost performance across multiple MIL frameworks. While Subsampling, DTFD-MIL, and ACMIL all show improvements over standard ABMIL, integrating AEM further elevates their performance. With Subsampling, AEM delivers additional gains, especially in cases where subsampling alone had limited impact. For DTFD-MIL, AEM contributes ~2\% AUC improvements on C17 and LBC datasets across all backbones. Even when paired with ACMIL, which addresses similar attention concentration issues, AEM still provides notable enhancements on C16 and C17 datasets while maintaining comparable performance on LBC. These consistent improvements across different methods highlight AEM's versatility as a complementary enhancement for diverse MIL approaches.

\noindent\textbf{Performance gains of AEM across different attention mechanisms.}
To validate AEM's versatility beyond gated attention, we applied it to three additional mechanisms: DSMIL \cite{li2021dual}, MHA \cite{zhang2023attention}, and LongNet \cite{xu2024whole}. Figure \ref{fig:com} shows AEM's impact across datasets and feature extraction methods. For DSMIL, improvements are most significant with VLM features on C17/SSL and LBC datasets, though C16/SSL performs slightly better without AEM. MHA shows more modest benefits, particularly on C17 and LBC datasets, likely due to its inherent capacity for learning diverse attention values \cite{li2021dual,zhang2023attention}. With LongNet using pretrained Gigapath \cite{xu2024whole} checkpoints, AEM consistently improves finetuning results on both CAMELYON datasets. AEM's effectiveness varies by context, showing particular promise with DSMIL+VLM features, complex datasets, and LongNet architectures.

\begin{figure}[t]
\centering
\begin{subfigure}{0.49\columnwidth}
\includegraphics[width=\textwidth]{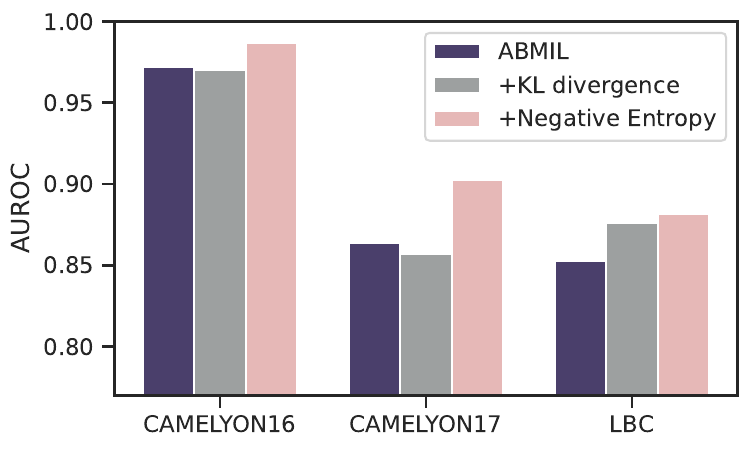}
\caption{Performance comparison of negative entropy vs. KL divergence as AEM loss formulation. Negative entropy shows greater stability and consistent gains.}
\label{fig:loss_formulation}
\end{subfigure}
\hfill
\begin{subfigure}{0.49\columnwidth}
\includegraphics[width=\textwidth]{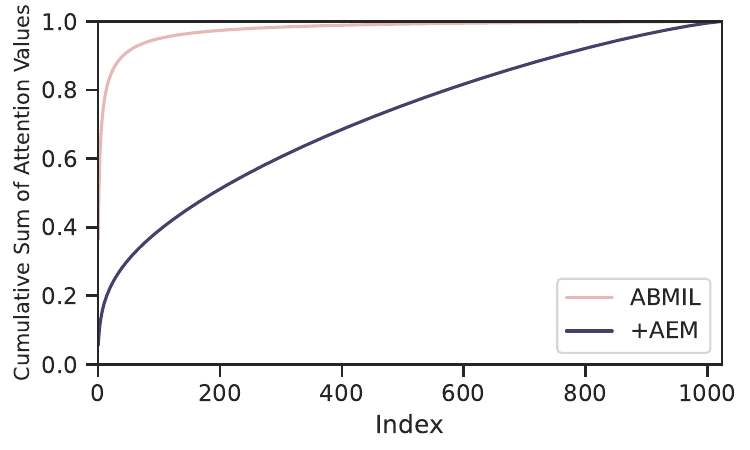}
\caption{Mean cumulative sum of top-1000 attention values for LBC test set. AEM effectively mitigates attention over-concentration.}
\label{fig:attention_distribution}
\end{subfigure}
\label{fig:combined_analysis}
\end{figure}

%setting $\lambda > 0$ consistently enhances performance compared to the baseline ($\lambda = 0$) on all three datasets. Moreover, different $\lambda$ should be set for different datasets to achieve the best performance. Specifically, $\lambda=0.005$ for C16, $\lambda=0.2$ for C17, and $\lambda=0.1$ for LBC. Thus, selecting an appropriate $\lambda$ is crucial for optimizing the effectiveness of AEM.

\begin{figure*}[t]
	\centering
	\begin{subfigure}[b]{0.24\textwidth}
		\includegraphics[width=\textwidth]{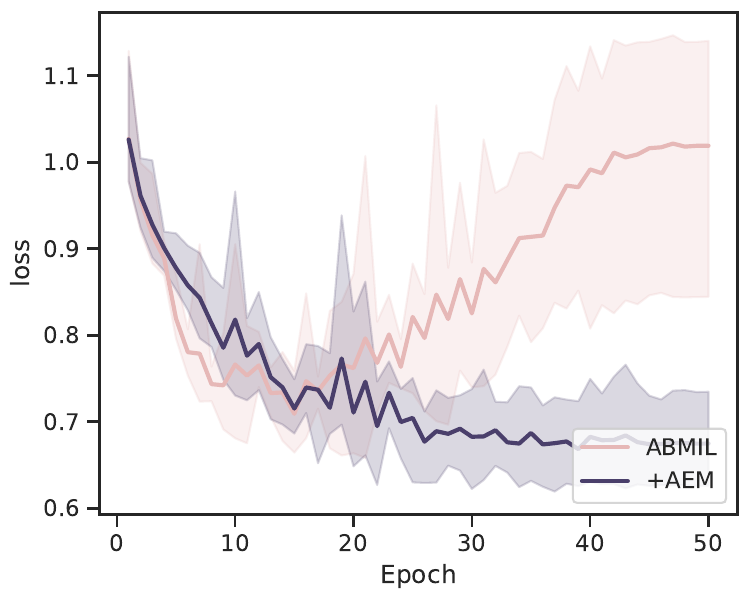}
		\caption{Test Loss}
		\label{fig:lct_test_loss}
	\end{subfigure}
	\begin{subfigure}[b]{0.24\textwidth}
		\includegraphics[width=\textwidth]{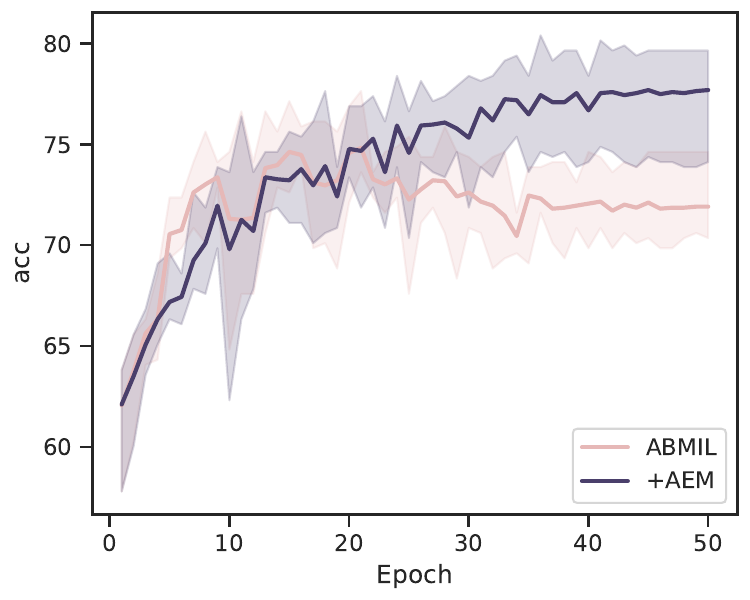}
		\caption{Test Accuracy}
		\label{fig:lct_test_acc}
	\end{subfigure}
	\begin{subfigure}[b]{0.24\textwidth}
		\includegraphics[width=\textwidth]{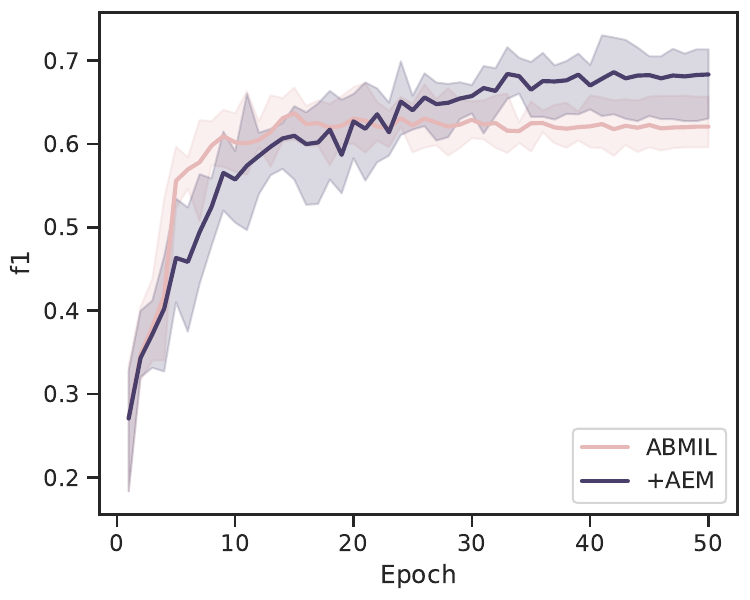}
		\caption{Test F1-score}
		\label{fig:lct_test_f1}
	\end{subfigure}
	\begin{subfigure}[b]{0.24\textwidth}
		\includegraphics[width=\textwidth]{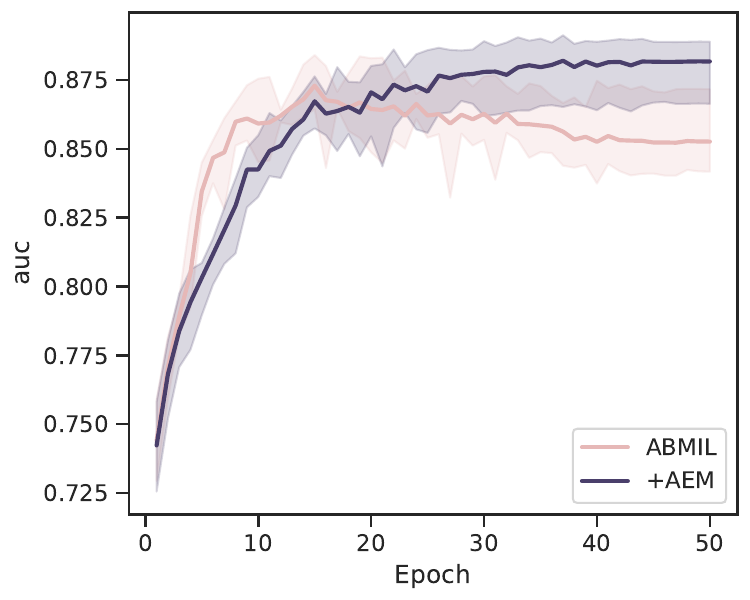}
		\caption{Test AUROC}
		\label{fig:lct_test_auc}
	\end{subfigure}
	\caption{Performance comparison between ABMIL \cite{ilse2018attention} and our AEM on LBC test set throughout training. \textit{ABMIL shows clear overfitting with increasing test loss and declining metrics, while AEM effectively prevents this issue.}}\label{fig:valid}
\end{figure*}

\subsection{Further analysis}

\noindent\textbf{Ablation Study.}
We examined the role of $\lambda$ across three datasets, testing values $\{0, 0.001, 0.002, 0.005, 0.01, 0.02, 0.05\}$ on C16, and $\{0, 0.01, 0.02, 0.05, 0.1, 0.2, 0.5\}$ on C17 and LBC, with $\lambda = 0$ representing baseline ABMIL. Figure \ref{fig:sensitivity} reveals that: 1) optimal $\lambda$ values are approximately 0.01 for C16 and 0.2 for C17/LBC; 2) CWA substantially improves stability, especially at higher $\lambda$ values where AEM without CWA degrades; 3) both AEM variants outperform the ABMIL baseline; and 4) CWA enables effective operation at larger $\lambda$ values by adaptively modulating attention weights during training.

\noindent\textbf{Superiority of Negative Entropy over KL Divergence.}
Comparing loss formulations for alleviating attention concentration, Figure \ref{fig:loss_formulation} shows AUROC results across three datasets using VLM pretrained embeddings. The negative entropy consistently outperformed both ABMIL baseline and KL divergence approaches. While KL divergence improved results on LBC, it degraded performance on CAMELYON datasets. Negative entropy provides more consistent and stable improvements, making it the preferred formulation for AEM.

\noindent\textbf{AEM effectively mitigates the overfitting.} Figure \ref{fig:valid} reveals that AEM maintains lower test loss, higher accuracy, and superior F1-score and AUROC compared to ABMIL across training epochs, with ABMIL showing signs of overfitting after epoch 20-30. AEM's consistent outperformance across all metrics demonstrates its superior generalization ability and robustness, establishing it as a more reliable approach less susceptible to overfitting than ABMIL.

\noindent\textbf{AEM effectively mitigates the attention concentration.} 
Figure \ref{fig:attention_distribution} demonstrates how AEM effectively mitigates the attention concentration problem observed in ABMIL for the LBC test set. The ABMIL curve (purple) rises sharply, indicating that it focuses most of its attention on a small subset of patches. In contrast, the AEM curve (brown) shows a much more gradual increase, suggesting a more balanced distribution of attention across a larger number of patches.

% \noindent\textbf{Additional results.} More results are presented in the Supplementary Material, including evaluation metric curves to visualize heatmap visualizations to demonstrate improved interpretability, UMAP visualizations of bag features to show that AEM learns more discriminative representations, and a detailed analysis of integrating AEM with DTFD-MIL, ACMIL, and Subsampling. 

\section{Conclusion}

This paper introduces AEM, a novel approach addressing attention concentration and overfitting in MIL frameworks through negative entropy regularization of instance attention distributions. AEM effectively mitigates these issues while offering advantages in simplicity—requiring no additional modules or processing. Our experiments demonstrate AEM enhances performance when combined with various MIL frameworks, attention mechanisms, and feature extractors, positioning it as a versatile enhancement for medical image analysis.

\noindent\textbf{Limitation and future work.} While currently focused on WSI classification, future work will extend AEM to survival and mutation prediction tasks. Though we introduced cosine weight annealing to stabilize training, the initial weight parameter still requires manual tuning. Future research will develop automatic weight adjustment mechanisms and investigate the theoretical bounds of entropy-based attention regularization.

\noindent\textbf{Acknowledgements.}
This study was partially supported by the National Natural Science Foundation of China (Grant No. 92270108), Zhejiang Provincial Natural Science Foundation of China (Grant No. XHD23F0201), and the Research Center for Industries of the Future (RCIF) at Westlake University.

{
    \bibliographystyle{splncs04}
    \bibliography{main}
}
\end{document}